\documentclass[11pt,a4paper]{article}
\usepackage[hyphens]{url}
\usepackage[hyphenbreaks]{breakurl}
\usepackage{acl2021}

\usepackage{times}
\usepackage{latexsym}

\usepackage{array}
\usepackage{booktabs}
\usepackage{amsmath}
\usepackage{amsfonts} % blackboard math symbols
\usepackage{xcolor}
\usepackage{color, colortbl}
\usepackage{adjustbox}
\usepackage{xspace}

\usepackage{arydshln}

\usepackage{CJKutf8}

\newcommand{\stitle}[1]{\vspace{0.3ex} \noindent{\bf #1}}

\usepackage{cleveref}
\crefformat{section}{\S#2#1#3}
\crefformat{subsection}{\S#2#1#3}
\crefformat{subsubsection}{\S#2#1#3}
\crefrangeformat{section}{\S\S#3#1#4 to~#5#2#6}
\crefmultiformat{section}{\S\S#2#1#3}{ and~#2#1#3}{, #2#1#3}{ and~#2#1#3}
\usepackage{refstyle}
\Crefformat{figure}{#2Fig.~#1#3}
\Crefmultiformat{figure}{Figs.~#2#1#3}{ and~#2#1#3}{, #2#1#3}{ and~#2#1#3}
\Crefformat{table}{#2Table~#1#3}
\Crefmultiformat{table}{Tabs.~#2#1#3}{ and~#2#1#3}{, #2#1#3}{ and~#2#1#3}
\Crefformat{equation}{#2Eq.~(#1#3)}

\Crefformat{appendix}{#2App.~\S#1#3}

\aclfinalcopy % Uncomment this line for the final submission
\definecolor{flc}{RGB}{255, 0, 255}
\definecolor{ncc}{RGB}{0, 153, 51}

\newcommand{\en}{{\textsc{en}}\xspace}
\newcommand{\de}{{\textsc{de}}\xspace}
\newcommand{\es}{{\textsc{es}}\xspace}
\newcommand{\ja}{{\textsc{ja}}\xspace}
\newcommand{\ko}{{\textsc{ko}}\xspace}
\newcommand{\zh}{{\textsc{zh}}\xspace}
\newcommand{\tha}{{\textsc{th}}\xspace}
\newcommand{\tr}{{\textsc{tr}}\xspace}
\newcommand{\fin}{{\textsc{fi}}\xspace}
\newcommand{\ru}{{\textsc{ru}}\xspace}
\newcommand{\fr}{{\textsc{fr}}\xspace}
\newcommand{\pt}{{\textsc{pt}}\xspace}
\newcommand{\nl}{{\textsc{nl}}\xspace}
\newcommand{\ita}{{\textsc{it}}\xspace}
\newcommand{\cs}{{\textsc{cs}}\xspace}
\newcommand{\et}{{\textsc{et}}\xspace}
\newcommand{\pl}{{\textsc{pl}}\xspace}
\newcommand{\no}{{\textsc{no}}\xspace}
\newcommand{\sv}{{\textsc{sv}}\xspace}
\newcommand{\hr}{{\textsc{hr}}\xspace}
\newcommand{\el}{{\textsc{el}}\xspace}
\newcommand{\lv}{{\textsc{lv}}\xspace}

\newcommand{\bel}{{\textsc{bel}}\xspace}
\newcommand{\xlbel}{{\textsc{xl-bel}}\xspace}
\newcommand{\bert}{{\textsc{Bert}}\xspace}
\newcommand{\sapbert}{{\textsc{SapBert}}\xspace}
\newcommand{\mbert}{{\textsc{mBert}}\xspace}
\newcommand{\xlmr}{{\textsc{XLMR}}\xspace}

\title{Learning Domain-Specialised Representations for Cross-Lingual Biomedical Entity Linking}

\author{Fangyu Liu, Ivan Vuli\'{c}, Anna Korhonen, Nigel Collier \\
Language Technology Lab, TAL, University of Cambridge\\
\texttt{\{fl399, iv250, alk23, nhc30\}cam.ac.uk}}

\date{}

\begin{document}
\maketitle
\begin{abstract}%   <- trailing '%' for backward compatibility of .sty file
%\iffalse
%In this paper, we highlight the challenge of transferring domain-specific knowledge in resource-rich languages to resource-poor ones. Specifically, we start with a state-of-the-art English entity representations model in the medical domain \textsc{SapBert} and demonstrate that general-domain bitext can help to propagate the stored English knowledge to languages with little to no in-domain data. To evaluate the effectiveness of our approach, we establish a medical entity linking benchmark covering 10 typologically and geographically diverse languages. We show that through our proposed transfer scheme, cross-lingual models can be enhanced on the task by as much as 20\% on Precision$_{@1}$ without relying on any in-domain parallel data.
%\fi

Injecting external domain-specific knowledge (e.g., UMLS) into pretrained language models (LMs) advances their capability to handle specialised in-domain tasks such as biomedical entity linking (\textsc{bel}). However, such abundant expert knowledge is available only for a handful of languages (e.g., English). In this work, by proposing a novel cross-lingual biomedical entity linking task (\xlbel) and establishing a new \xlbel benchmark spanning 10 typologically diverse languages, we first investigate the ability of standard knowledge-agnostic as well as knowledge-enhanced monolingual and multilingual LMs beyond the standard monolingual English \textsc{bel} task. The scores indicate large gaps to English performance. We then address the challenge of transferring domain-specific knowledge from resource-rich languages to resource-poor ones. To this end, we propose and evaluate a series of cross-lingual transfer methods for the \xlbel task, and demonstrate that general-domain bitext helps propagate the available English knowledge to languages with little to no in-domain data. Remarkably, we show that our proposed domain-specific transfer methods yield consistent gains across all target languages, sometimes up to 20 Precision${@1}$ points, without any in-domain knowledge in the target language, and without any in-domain parallel data.

%\footnote{For code, data, and pretrained models, please visit: \url{https://github.com/cambridgeltl/sapbert}.}

\end{abstract}

\section{Introduction}\label{sec:intro}
 Recent work has demonstrated that it is possible to combine the strength of 1) Transformer-based encoders such as \textsc{Bert} \cite{devlin2019bert,liu2019roberta}, pretrained on large general-domain data with 2) external linguistic and world knowledge \cite{Zhang:2019acl,Levine:2020acl,Lauscher:2020coling}. Such expert human-curated knowledge is crucial for NLP applications in specialised domains such as biomedicine. There, \citet{liu2020self} recently proposed \textit{self-alignment pretraining} (\textsc{Sap}), a technique to fine-tune \textsc{Bert} on phrase-level synonyms extracted from the Unified Medical Language System (UMLS; \citealt{bodenreider2004unified}).\footnote{UMLS is a large-scale biomedical knowledge graph containing more than 14M biomedical entity names.} Their \textsc{SapBert} model currently holds state-of-the-art (SotA) across all major English biomedical entity linking (\textsc{bel}) datasets. However, this approach is not widely applicable to other languages: abundant external resources are available only for a few languages, hindering the development of domain-specific NLP models in all other languages.
 
 %% (IV, redundant)
 %whether such knowledge can be adapted to other languages, especially the ones lacking learning resources. 

Simultaneously, exciting breakthroughs in cross-lingual transfer for language understanding tasks have been achieved \citep{artetxe2019massively,Hu:2020icml}. However, it remains unclear whether such transfer techniques can be used to improve domain-specific NLP applications and mitigate the gap between knowledge-enhanced models in resource-rich versus resource-poor languages. In this paper, we thus investigate the current performance gaps in the \bel task beyond English, and propose several cross-lingual transfer techniques to improve domain-specialised representations and \bel in resource-lean languages.

%In the meanwhile, exciting breakthroughs have been achieved in cross-lingual transfer learning leveraging parallel data. Using word-, phrase- or sentence-level bitext, prior works have reported enhanced performance on cross-lingual language understanding tasks \citep{artetxe2019massively,reimers-gurevych-2020-making}. That said, it is unknown whether such success can be extended to a specific domain, where language-use is significantly different from the general domain (e.g., the vocabulary in the scientific domain is much larger and more complex than the general domain).

In particular, we first present a novel cross-lingual \bel (\xlbel) task and its corresponding evaluation benchmark in 10 typologically diverse languages, which aims to map biomedical names/mentions in any language to the controlled UMLS vocabulary. After empirically highlighting the deficiencies of multilingual encoders (e.g, \mbert and \xlmr; \citealt{conneau2020unsupervised}) on \xlbel,  we propose and evaluate a multilingual extension of the \textsc{Sap} technique. Our main results suggest that expert knowledge can be transferred from English to resource-leaner languages, yielding huge gains over vanilla \mbert and \xlmr, and English-only \sapbert. We also show that leveraging general-domain word and phrase translations offers substantial gains in the \xlbel task.

%%In this paper, we initiate an investigation into bridging the gap of cross-lingual and domain-specific NLP research. Specifically, we enhance cross-lingual representations for biomedical entities and test a wide range of models' performances on our proposed task of cross-lingual biomedical entity linking (\xlbel). The task of \xlbel aims to map a biomedical mention in any language to the a controlled vocabulary (i.e., UMLS). The \xlbel benchmark is extracted from Wikipedia and covers 10 diverse languages: English (en), Spanish (es), German (de), Finnish (fi), Russian (ru), Korean (ko), Chinese (zh), Japanese (ja) and Thai th). We show that off-the-shelf SOTA cross-lingual pretrained language models such as \textsc{mBert} \citep{devlin2019bert} and \textsc{XLMR} \citep{conneau2020unsupervised} fail significantly on this benchmark. At the same time, existing SOTA domain-specific models such as \textsc{SapBert} \citep{liu2020self} are usually only available in English. We propose a multilingual extension of the Self-alignment pretraining (\textsc{Sap}) procedure \citep{liu2020self}. Moreover, our method can effectively leverage general domain word/phrase translations to boost performance by very large margins.

\vspace{1.6mm}
\noindent \textbf{Contributions.} \textbf{1)} We highlight the challenge of learning (biomedical) domain-specialised cross-lingual representations. \textbf{2)} We propose a novel multilingual \xlbel task with a comprehensive evaluation benchmark in 10 languages. \textbf{3)} We offer systematic evaluations of existing knowledge-agnostic and knowledge-enhanced monolingual and multilingual LMs in the \xlbel task. \textbf{4)} We present a new SotA multilingual encoder in the biomedical domain, which yields large gains in \xlbel especially on resource-poor languages, and provides strong benchmarking results to guide future work. The code, data, and pretrained models are available online at: \\\url{https://github.com/cambridgeltl/sapbert}.

\begin{table*}[t]
\small
\setlength{\tabcolsep}{1.8pt}
\centering
\begin{tabular}{lcccccccccccc}
\toprule
 %\multirow{2}{*}{model} &  \multicolumn{3}{c}{FR$\rightarrow$EN}  \\
 %\cmidrule{2-4}
\#$\downarrow$, language$\rightarrow$ & \en & \es & \de & \fin & \ru & \tr & \ko & \zh & \ja & \tha\\
\midrule
sentences & - & 223,506 & 350,193 & 77,736 & 206,060 & 29,473 & 47,702 & 136,054 & 157,670 & 19,066\\
unique titles (Wiki page) & 60,598 & 37,935 & 24,059 & 15,182 & 21,044 & 5,251 & 10,618 & 17,972 & 11,002 & 4,541 \\
mentions & 1,067,083 & 204,253 & 431,781 & 105,182 & 221,383 & 29,958 & 60,979 & 197,317 & 220,452 & 31,177 \\
unique mentions & 121,669 & 25,169 & 44,390 & 26,184 & 28,302 & 4,110 & 9,032 & 24,825 & 21,949 & 5,064 \\
unique mentions$_{\text{mention!=title}}$ & 69,199 & 22,162 & 43,753 &  19,409 & 23,935 & 2,833 & 3,740 & 12,046 & 12,571 & 2,480\\
\bottomrule
\end{tabular}
%\vspace{-0.4em}
\caption{Construction of the \xlbel benchmark; key statistics. See the \Cref{sec:appendix_xlbel} for further details.}
\label{tab:xlbel}
\end{table*}

\section{Methodology}\label{sec:method}

%\noindent \textbf{Background and Related Work.} 
\noindent \textbf{Background and Related Work.} 
Learning biomedical entity representations is at the core of BioNLP, benefiting, e.g., relational knowledge discovery \citep{wang2018comparison} and literature search \citep{lee2016best}. In the current era of contextualised representations based on Transformer architectures \cite{Vaswani:2017nips}, biomedical text encoders are pretrained via Masked Language Modelling (MLM) on diverse biomedical texts such as PubMed articles \citep{lee2020biobert,pubmedbert}, clinical notes \citep{peng2019transfer,clinicalbert}, and even online health forum posts \citep{cometa}. However, it has been empirically verified that naively applying MLM-pretrained models as entity encoders does not perform well in tasks such as biomedical entity linking \citep{cometa,sung-etal-2020-biomedical}. Recently, \citet{liu2020self} proposed \textsc{Sap} (\textbf{S}elf-\textbf{A}lignment \textbf{P}retraning), a fine-tuning method that leverages synonymy sets extracted from UMLS to improve \bert's ability to act as a biomedical entity encoder. Their \sapbert model currently achieves SotA scores on all major English \textsc{bel} benchmarks. 

In what follows, we first outline the \textsc{Sap} procedure, and then discuss the extension of the method to include multilingual UMLS synonyms (\Cref{sec:lasap}), and then introduce another \textsc{Sap} extension which combines domain-specific synonyms with general-domain translation data (\Cref{sec:sap_bitext}).

%into class-based data, thus \textsc{Sap} can be directly applied for learning from bitext (\Cref{sec:sap_bitext}).

\subsection{Language-Agnostic \textsc{Sap}}\label{sec:lasap}

Let $(x, y)\in \mathcal{X} \times \mathcal{Y}$ denote the tuple of a name and its categorical label.
When learning from UMLS synonyms, $\mathcal{X} \times \mathcal{Y}$ is the set of all \textit{(name, CUI\footnote{In UMLS, ``CUI'' means \textbf{C}oncept \textbf{U}nique \textbf{I}dentifier.}}) pairs, e.g., (\emph{vaccination}, \texttt{C0042196}). While \citet{liu2020self} use only English names, we here consider names in other UMLS languages. During training, the model is steered to create similar representations for synonyms regardless of their language.\footnote{For instance, \emph{vaccination} (\en), \emph{active immunization} (\en), \emph{vacunación} (\es) and \begin{CJK}{UTF8}{min}\emph{予防接種}\end{CJK} (\ja) all share the same Concept Unique Identifier (CUI; \texttt{C0042196}); thus, they should all have similar representations.} The learning scheme includes 1) an online sampling procedure to select training examples and 2) a metric learning loss that encourages strings sharing the same CUI to obtain similar representations. %live close to each other in the embedding space (both introduced below).

%\vspace{1.2mm}
%\noindent \textbf{Training Examples.} 
\paragraph{Training Examples.}
Given a mini-batch of $N$ examples $\mathcal{B}= \mathcal{X_B}\times \mathcal{Y_B} = \{(x_i, y_i)\}_{i=1}^{N}$, we start from constructing all possible triplets for all names $x_i \in \mathcal{X_B}$. Each triplet is in the form of $(x_a, x_p, x_n)$ where $x_a$ is the \emph{anchor}, an arbitrary name from $\mathcal{X_B}$; $x_p$ is a positive match of $x_a$ (i.e., $y_a=y_p$) and $x_n$ is a negative match of $x_a$ (i.e., $y_a \neq y_n$). Let $f(\cdot)$ denote the encoder (i.e., \mbert or \xlmr in this paper). Among the constructed triplets, we select all triplets that satisfy the following constraint:
\begin{equation}
    \|f(x_a)-f(x_p)\|_2 + \lambda \ge \|f(x_a)-f(x_n)\|_2 , \notag
    \label{eq:mining}
\end{equation}
where $\lambda$ is a predefined margin. 
In other words, we only consider triplets with the positive sample further to the negative sample by a margin of $\lambda$. These `hard' triplets are more informative for representation learning \citep{liu2020self}.
%We call them 
Every selected triplet then contributes one positive pair $(x_a,x_p)$ and one negative pair $(x_a,x_n)$. We collect all such positives and negatives, and denote them as $\mathcal{P},\mathcal{N}$. 

%% a SOTA metric learning objective on visual recognition, 

%\vspace{1.2mm}
%\noindent \textbf{Multi-Similarity Loss.} 
\paragraph{Multi-Similarity Loss.}
We compute the pairwise cosine similarity of all the name representations and obtain a similarity matrix $\mathbf{S}\in\mathbb{R}^{|\mathcal{X_B}|\times |\mathcal{X_B}|}$ where each entry $\mathbf{S}_{ij}$ is the cosine similarity between the $i$-th and $j$-th names in the mini-batch $\mathcal{B}$. The Multi-Similarity loss~(MS, \citealt{wang2019multi}), is then used for learning from the triplets:

%\vspace{-1.5mm}
{\footnotesize
\begin{equation}
    \begin{split}
\mathcal{L} = \frac{1}{|\mathcal{X}_\mathcal{B}|}\sum_{i=1}^{|\mathcal{X}_\mathcal{B}|} \Bigg(\frac{1}{\alpha}\log\Big(1+\sum_{n\in \mathcal{N}_i } e^{\alpha (\mathbf{S}_{in} -\epsilon )}\Big) \\
 + \frac{1}{\beta} \log \Big(1+\sum_{p \in \mathcal{P}_{i}} e ^{ - \beta (\mathbf{S}_{ip} -\epsilon)} \Big)\Bigg) .
\label{eq:loss}
\end{split}
\end{equation}}%
\noindent $\alpha,\beta$ are temperature scales; $\epsilon$ is an offset applied on the similarity matrix; $\mathcal{P}_i,\mathcal{N}_i$ are indices of positive and negative samples of the $i$-th \emph{anchor}.

%%While the first term in Eq.~\ref{eq:loss} pushes negative pairs away from each other, the second term pulls positive pairs together. 
%This dynamic allows for a  re-calibration of the alignment space using the semantic biases of synonymy relations.

\subsection{\textsc{Sap} with General-Domain Bitext}\label{sec:sap_bitext}

%% Namely, the two strings in the bitext pair shares the same categorical label and the label of different bitext pairs are different. 

%% In this way, the cross-lingual bitext can be directly used for \textsc{Sap}. 

We also convert word and phrase translations into the same format (\S\ref{sec:lasap}), where each `class' now contains only two examples. For a translation pair $(x_p, x_q)$, we create a unique pseudo-label $y_{x_p,x_q}$ and produce two new name-label instances $(x_p, y_{x_p,x_q})$ and $(x_q, y_{x_p,x_q})$,\footnote{These pseudo-labels are not related to UMLS, but are used to format our parallel translation data into the input convenient for the \textsc{Sap} procedure. In practice, for these data we generate pseudo-labels ourselves as `LANGUAGE\_CODE+index'. For instance, \texttt{ENDE2344} indicates that this word pair is our 2,344th English-German word translation. Note that the actual coding scheme does not matter as it is only used for our algorithm to determine what terms belong to the same (in this case - translation) category.} and proceed as in \S\ref{sec:lasap}. This allows us to easily combine domain-specific knowledge with general translation knowledge within the same \textsc{Sap} framework.

%%In experiments, we will show that by first conducting \textsc{Sap} on UMLS data (as described in \Cref{sec:lasap}), and then further finetuned on general-domain cross-lingual bitext, LMs become powerful cross-lingual domain experts, showing promising performance for resource-low languages.

\section{The \xlbel Task and Evaluation Data}
\label{sec:xlbel}

A general cross-lingual entity linking (EL) task \citep{mcnamee2011cross,tsai2016cross} aims to map a mention of an entity in free text of \textit{any language} to a controlled English vocabulary, typically obtained from a knowledge graph (KG). In this work, we propose \xlbel, a cross-lingual \textit{biomedical} EL task. Instead of grounding entity mentions to English-specific ontologies, we use UMLS as a language-agnostic KG: the \xlbel task requires a model to associate a mention in any language to a (language-agnostic) CUI in UMLS. \xlbel thus serves as an ideal evaluation benchmark for biomedical entity representations: it challenges the capability of both 1) representing domain entities and also 2) associating entity names in different languages.

%\vspace{1.3mm}
%\noindent \textbf{Evaluation Data Creation.}
\paragraph{Evaluation Data Creation.}
%, addressing \textsc{xlel} in the biomedical domain.
%Cross-lingual entity linking (\textsc{xlel}) aims to ground a mention \emph{in any language} to an English KG \citep{mcnamee2011cross,tsai2016cross}. 
%In this work, we propose the \xlbel task, addressing \textsc{xlel} in the biomedical domain.  While \textsc{xlel} maps multilingual mentions to a language-specific ontology (English), we employ a language-agnostic KG (UMLS) as the target labels for all languages (though the distribution of synonyms is heavily skewed towards English, see \Cref{tab:umls}). 
%The \xlbel task requires a model to associate a mention in any language to a CUI in UMLS. 
For English, we take the available \textsc{bel} dataset WikiMed \citep{medtype2020}, which links Wikipedia mentions to UMLS CUIs. We then follow similar procedures as WikiMed and create an \xlbel benchmark covering 10 languages (see \Cref{tab:model_sap}). For each language, we extract all sentences from its Wikipedia dump, find all hyperlinked concepts (i.e., words and phrases), lookup their Wikipedia pages, and retain only concepts that are linked to UMLS.\footnote{For instance, given a sentence from German Wikipedia \emph{Die [Inkubationszeit] von COVID-19 beträgt durchschnittlich fünf bis sechs Tage.}, we extract the hyperlinked word \emph{Inkubationszeit} as an UMLS-linked entity mention. Since Wikipedia is inherently multilingual, if \emph{Inkubationszeit} is linked to UMLS, its cross-lingual counterparts, e.g., \emph{Incubation period} (\en), are all transitively linked to UMLS.} For each UMLS-linked mention, we add a triplet \textit{(sentence, mention, CUI)} to our dataset.\footnote{Note that though each mention is accompanied with its context, we regard it as out-of-context mention following the tradition in prior work \citep{sung-etal-2020-biomedical,liu2020self,tutubalina-etal-2020-fair}. 
According to \citet{cometa}, biomedical entity representations can be easily polluted by its context. We leave contextual modelling for future work.} 
Only one example per surface form is retained to ensure diversity. 
We then filter out examples with mentions that have the same surface form as their Wikipedia article page.\footnote{Otherwise, the problem is easily solved by comparing surface forms of the mention and the article title.}
Finally, 1k examples are randomly selected for each language: they serve as the final test sets in our \xlbel benchmark. The statistics of the benchmark are available in \Cref{tab:xlbel}.

\begin{table*}[!t]
\scriptsize
\setlength{\tabcolsep}{1.8pt}
%\vspace*{-0.3cm}
%\renewcommand{\arraystretch}{0.9}
%\vspace{-0.5em}
%\vspace{-0.5em}
\centering
\begin{tabular}{llccccccccccccccccccccccccccccccccc}
\toprule
  language$\rightarrow$& \multicolumn{2}{c}{\en} & $\ $ & \multicolumn{2}{c}{\es} & $\ $ & \multicolumn{2}{c}{\de} & $\ $ &  \multicolumn{2}{c}{\fin} & $\ $ &  \multicolumn{2}{c}{\ru} & $\ $ & 
  \multicolumn{2}{c}{\tr} & $\ $ &
  \multicolumn{2}{c}{\ko} & $\ $ & \multicolumn{2}{c}{\zh} & $\ $ &  \multicolumn{2}{c}{\ja} &  $\ $ &  \multicolumn{2}{c}{\tha} & $\ $ &  \multicolumn{2}{c}{\bf avg}\\
 \cmidrule{2-3}\cmidrule{5-6} \cmidrule{8-9} \cmidrule{11-12} \cmidrule{14-15} \cmidrule{17-18}\cmidrule{20-21}\cmidrule{23-24}\cmidrule{26-27}\cmidrule{29-30}\cmidrule{32-33}
  model$\downarrow$ &  {\scriptsize @1} &\scriptsize @5 & &\scriptsize @1 &\scriptsize @5 & &\scriptsize @1 &\scriptsize @5 & &\scriptsize @1 &\scriptsize @5 & &\scriptsize @1 &\scriptsize @5 & &\scriptsize @1 &\scriptsize @5 & &\scriptsize @1 &\scriptsize @5 & &\scriptsize @1 &\scriptsize @5 & &\scriptsize @1 &\scriptsize @5 & &\scriptsize @1 &\scriptsize @5  & &\scriptsize @1 &\scriptsize @5  \\
  \midrule
 \multicolumn{6}{l}{\xspace\xspace\emph{monolingual models}}   \\
  \midrule
   \rowcolor{black!10}
    \{\$LANG\}\textsc{Bert} & - & - && 41.3 & 42.5 && 16.8 & 18.4 && 4.9 & 5.2 && 1.1 & 1.6 && 19.5 & 21.8 && 1.1 & 1.6 && 2.1 & 3.2 && 2.7 & 2.8 && 0.4 & 0.4 && 10.0 & 10.8  \\  
    %\{\$LANG\}\textsc{Bert}$_{\textsc{nospec}}$ & - & - && 42.1 & 44.2 && 9.5 & 12.1 && 2.4 & 3.3 && 2.4 & 3.5 && 24.1 & 25.8 && 0.1 & 0.4 && 1.5 & 1.9 && 1.9 & 2.5 && 0.5 & 0.5 && 9.4 & 10.5 \\
     \rowcolor{cyan!10}
 \ \ \ \textsc{+ Sap}$_{\text{all\_syn}}$ & - & - && 60.9 & 66.8 && \textbf{35.5} & \textbf{40.0} && \textbf{18.8} & \textbf{23.9} && \textbf{36.4} & \textbf{42.4} && \textbf{44.9} & \textbf{49.7} && 13.5 & 16.0 && 18.5 & \textbf{23.8} && 21.2 & 25.9 && 0.6 & 0.6 && 27.8 & 32.1 \\
    \hline
    \rowcolor{cyan!10}
% \citep{liu2020self}} 
  \textsc{SapBert} & \textbf{78.7} & \textbf{81.6} && 47.3 & 51.4 && 22.7 & 24.7 && 8.2 & 10.2 && 5.8 & 6.0 && 26.4 & 29.7 && 2.0 & 2.4 && 1.9 & 2.2 && 3.0 & 3.2 && 3.1 & 3.4 && 19.9 & 21.6 \\
    \rowcolor{cyan!10}
    \textsc{SapBert}$_{\text{all\_syn}}$ & 78.3 & 80.7 && 55.6 & 61.3 && 30.0 & 34.2 && 11.8 & 14.8 && 9.3 & 11.3 && 35.5 & 39.5 && 2.0 & 2.4 && 6.4 & 8.2 && 6.9 & 8.3 && 3.0 & 3.3 && 23.9 & 26.4 \\
    \midrule
 \multicolumn{6}{l}{\xspace\xspace\emph{multilingual models}}   \\
  \midrule
    \rowcolor{black!10}
\textsc{mBert} & 0.8 & 1.7 &&  0.5 & 0.7 && 0.3 & 0.4 && 0.4 & 0.8 && 0.0 & 0.0 && 0.7 & 1.2 && 0.0 & 0.0 &&  0.0 & 0.0 && 0.0 & 0.1 && 0.0 & 0.0 && 0.3 & 0.5 \\
  %\textsc{mBert}$_{\textsc{nospec}}$ & 35.8  & 49.4 && 15.0 & 24.1 && 7.1 & 10.4 && 1.7 & 2.6 && 2.3 & 3.6 && 10.8 & 15.3 &&  1.3 & 1.7 && 0.9 & 1.4 && 1.7 & 2.2 && 1.1 & 1.7 && 7.8 & 11.2 \\
  % \textsc{mBert-Large}$_{\textsc{nospec}}$ & & &&\\
    \rowcolor{cyan!10}
\ \ \ \textsc{+ Sap}$_{\text{en\_syn}}$ & 75.5 & 79.9 && 50.6 & 55.8 && 26.0 & 29.6 && 8.7 & 10.7 && 10.1 & 12.6 && 31.0 & 34.4 && 2.7 & 3.2 && 4.1 & 5.7 && 4.7 & 5.9 && 3.1 & 3.5 && 21.7 & 24.1 \\
  \rowcolor{cyan!10}
   \ \ \ \textsc{+ Sap}$_{\text{all\_syn}}$ & 75.0 & 79.7 && \textbf{61.4} & \textbf{67.0} &&  33.4 & 37.8 && 18.4 & 21.9 && 35.1 & 40.3 && 44.5 & 47.7 && 15.1 & 17.6 && \textbf{19.5} & 22.7 && 19.9 & 25.0 && 2.8 & 3.4 && 32.5 & 36.3 \\
     \hline  
     \rowcolor{black!10}
\xlmr & 1.0 & 2.0 && 0.3 & 0.7 && 0.0 & 0.1 && 0.1 & 0.2 && 0.1 & 0.2 && 0.4 & 0.5 && 0.0 & 0.3 && 0.1 & 0.2 && 0.2 & 0.4 && 0.0 & 0.1 && 0.2 & 0.5 \\
%\textsc{XLMR}$_{\textsc{nospec}}$ & 0.2 & 0.4 && 0.0 & 0.0 && 0.2 & 0.3 && 0.1 & 0.2 && 0.2 & 0.2 && 0.0 & 0.0 && 0.0 & 0.2 && 0.1 & 0.1 && 0.0 & 0.2 && 0.1 & 0.3 && 0.1 & 0.2 \\
%\textsc{XLMR} & 0.2 & 0.4 && 0.0 & 0.0 && 0.2 & 0.3 && 0.1 & 0.2 && 0.2 & 0.2 && 0.0 & 0.0 && 0.0 & 0.2 && 0.1 & 0.1 && 0.0 & 0.2 && 0.1 & 0.3 && 0.1 & 0.2 \\
%\textsc{XLMR-Large} & 73.0 & 75.0 && 20.7 & 24.6 && 7.8 & 9.1 && 1.9 & 2.7 && 3.0 & 3.3 && 11.8 & 13.5 && 1.2 & 1.2 && 0.7 & 0.9 && 1.6 & 1.8 && 0.9 & 1.2\\
%\textsc{XLMR-Large}$_{\textsc{nospec}}$ & 27.0 &  29.8 && 10.5 & 13.1 && 3.7 & 4.4 && 1.3 & 1.6 \\
  \rowcolor{cyan!10}
\ \ \ \textsc{+ Sap}$_{\text{en\_syn}}$ & 78.1 & 80.9 && 47.9 & 53.5 &&  27.6 & 32.0 && 12.2 & 14.7 && 21.8 & 25.9 && 29.3 & 35.9 && 4.5 & 6.7 && 7.9 & 11.3 && 8.3 & 11.3 && 11.5 & 16.2 && 24.9 & 28.8 \\
  \rowcolor{cyan!10}
\ \ \ \textsc{+ Sap}$_{\text{all\_syn}}$ & 78.2 & 81.0 && 56.4 & 62.7 & & 31.8 & 37.3 && 18.6 & 22.2 && 35.4 & 41.2 && 42.8 & 48.9 && \textbf{16.7} & \textbf{21.4} && 18.8 & 23.0 && \textbf{24.0} & \textbf{28.1} && \textbf{20.6} & \textbf{27.5} && \textbf{34.3} & \textbf{39.3} \\
\bottomrule
\end{tabular}
%\vspace{-1mm}
\caption{Various base models combined with \textsc{Sap}, using either all synonyms (\textit{all\_syn}) or only English synonyms (\textit{en\_syn}) in UMLS. \{\$LANG\} denotes the language of the corresponding column (also in \Cref{tab:the_more_the_better}). See \Cref{tab:umls} (\Cref{sec:appendix_umls}) for the language codes. \textbf{avg} refers to the average performance across all target languages. \colorbox{black!10}{Grey} and \colorbox{cyan!10}{light blue} rows are off-the-shelf base models and models fine-tuned with the UMLS knowledge, respectively.}
%\vspace{-1.5mm}
\label{tab:model_sap}
\end{table*}
\begin{table*}[!t]
\scriptsize
\setlength{\tabcolsep}{1.8pt}
%\def\arraystretch{0.75}
%\vspace*{-0.3cm}
%\renewcommand{\arraystretch}{0.9}
%\vspace{-0.5em}
%\vspace{-0.5em}
\centering
\resizebox{1.0\textwidth}{!}{
\begin{tabular}{llccccccccccccccccccccccccccccc}
\toprule
  language$\rightarrow$& \multicolumn{2}{c}{\es} & $\ $ & \multicolumn{2}{c}{\de} & $\ $ &  \multicolumn{2}{c}{\fin} & $\ $ &  \multicolumn{2}{c}{\ru} & $\ $ & 
  \multicolumn{2}{c}{\tr} & $\ $ &
  \multicolumn{2}{c}{\ko} & $\ $ & \multicolumn{2}{c}{\zh} & $\ $ &  \multicolumn{2}{c}{\ja} &  $\ $ &  \multicolumn{2}{c}{\tha} &  $\ $ &  \multicolumn{2}{c}{\bf avg}\\
 \cmidrule{2-3}\cmidrule{5-6} \cmidrule{8-9} \cmidrule{11-12} \cmidrule{14-15} \cmidrule{17-18}\cmidrule{20-21}\cmidrule{23-24}\cmidrule{26-27}\cmidrule{29-30}
  model$\downarrow$ &  {\scriptsize @1} &\scriptsize @5 & &\scriptsize @1 &\scriptsize @5 & &\scriptsize @1 &\scriptsize @5 & &\scriptsize @1 &\scriptsize @5 & &\scriptsize @1 &\scriptsize @5 & &\scriptsize @1 &\scriptsize @5 & &\scriptsize @1 &\scriptsize @5 & &\scriptsize @1 &\scriptsize @5 & &\scriptsize @1 &\scriptsize @5 & &\scriptsize @1 &\scriptsize @5 &\\
  \midrule
  \rowcolor{cyan!10}
\textsc{XLMR + Sap}$_{\text{en\_syn}}$ & 47.9 & 53.5 &&  27.6 & 32.0 && 12.2 & 14.7 && 21.8 & 25.9 && 29.3 & 35.9 && 4.5 & 6.7 && 7.9 & 11.3 && 8.3 & 11.3 && 11.5 & 16.2 && 19.0 & 23.1 \\
 % \hdashline
 \rowcolor{magenta!10}
 \ \ \ + en-\{\$LANG\} wt & 55.0 & 62.2 & & 34.6 & 41.4 && 18.6 & 24.4 && 35.0 & 41.5 && 43.3 & 50.6 && 15.9 & 22.3 && 15.9 & 23.0 && 18.7 & 24.4 && 25.1 & 32.4 && 29.1 & 35.8 \\
  \rowcolor{magenta!10}
 \ \ \ + en-\{\$LANG\} muse & 54.4 & 61.0 && 28.7 & 34.4 && 16.7 & 20.6 && 33.6 & 39.0 && 41.9 & 48.8 && 11.9 & 16.3 && 12.3 & 16.7 && 15.7 & 19.9 && 18.6 & 25.1 && 26.0 & 31.3 \\
  \rowcolor{magenta!10}
 \ \ \ + en-\{\$LANG\} wt+muse & 49.4 & 59.6 &  & 30.3 & 36.9 && 20.4 & 28.9 && 33.2 & 41.9 && 42.7 & 51.7 &&  16.1 & 22.3 && 16.0 & 22.9 && 17.8 & 24.3 && 26.2 & 34.0 && 28.0 & 35.8 \\
  \hline  
  \rowcolor{cyan!10}
 \textsc{XLMR + Sap}$_{\text{all\_syn}}$ & 56.4 & 62.7 & & 31.8 & 37.3 && 18.6 & 22.2 && 35.4 & 41.2 && 42.8 & 48.9 && 16.7 & 21.4 && 18.8 & 23.0 && 24.0 & 28.1 && 20.6 & 27.5 && 29.5 & 34.7  \\
%  \hdashline
\rowcolor{magenta!10}
  \ \ \ + en-\{\$LANG\} wt & 57.2 & 63.7  && 35.1 & 42.3 && 20.3 & 27.6 && 35.8 & 43.8 && 48.8 & 55.0 && 22.1 & 27.9 && 20.6 & 27.3 && 24.8 & \textbf{31.3} && 30.0 & 37.6 && 32.7 & \textbf{39.6} \\
  \rowcolor{magenta!10}
  \ \ \ + en-\{\$LANG\} muse & 57.9 & 63.9 && 33.0 & 38.4 && 23.0 & 27.3 && 39.8 & 45.9 && 47.2 & 54.5 && 22.1 & 25.7 && 19.2 & 25.6 && \textbf{25.2} & 30.2 && 25.9 & 32.8 && 32.6 & 38.3  \\
  \rowcolor{magenta!10}
  \ \ \ + en-\{\$LANG\} wt+ muse & 51.4 & 61.2 && 31.3 & 38.9 && 22.8 & 28.4 && 36.4 & 45.2 && 42.2 & 51.6 && \textbf{24.4} & \textbf{29.2} && 21.1 & 28.2 && 23.2 & 30.4 && \textbf{30.9} &\textbf{37.9} && 31.5 & 39.0 \\
  \hline  
  \rowcolor{cyan!10}
 \textsc{mBert + Sap}$_{\text{all\_syn}}$ & \textbf{61.4} & 67.0 && 33.4 & 37.8 && 18.4 & 21.9 && 35.1 & 40.3 && 44.5 & 47.7 && 15.1 & 17.6 && 19.5 & 22.7 && 19.9 & 25.0 && 2.8 & 3.4 && 27.8 & 31.5 \\
%\hdashline
\rowcolor{magenta!10}
  \ \ \ + en-\{\$LANG\} wt & 59.2 & 66.9 && \textbf{37.5} & \textbf{43.9} && 25.6 & 33.0 && 39.6 & 47.2 && 52.7 & 59.7 && 19.8 & 24.3 && 24.1 & 31.9 && 23.5 & 28.7 && 4.8 & 5.9 && 31.9 & 37.9 \\
  \rowcolor{magenta!10}
 \ \ \ + en-\{\$LANG\} muse & 59.9 & 66.2 && 34.3 & 38.8 && 21.6 & 27.5 && 36.5 & 41.7 && 51.0 & 56.7 && 18.1 & 21.2 && 22.2 & 26.4 && 22.0 & 25.5 && 3.4 & 3.8 && 29.2 & 34.2 \\
 \rowcolor{magenta!10}
  \ \ \ + en-\{\$LANG\} wt+ muse & 59.2 & \textbf{67.5} && 35.3 & 42.4 && \textbf{30.5} & \textbf{37.3} && \textbf{41.6} & \textbf{49.2} && \textbf{57.2} & \textbf{64.7} && 19.8 & 25.0 && \textbf{24.6} & \textbf{32.1} && 24.3 & 28.0 && 5.2 & 6.3 &&  \textbf{33.1} & 39.2 \\
\bottomrule
\end{tabular}
}%
%\vspace{-1mm}
\caption{Results when applying \textsc{Sap} with \colorbox{cyan!10}{1) UMLS knowledge} + \colorbox{magenta!10}{2) word and/or phrase translations}.}
\label{tab:bitext_helps}
%\vspace{-0.6mm}
\end{table*}

\section{Experiments and Results}\label{sec:exp}
%(everything below is under construction...)
%% (IV, too fluff for the short paper)
%%We first introduce the data, training and evaluation setups in \Cref{sec:data_training_eval}; then we present and discuss the main results in \Cref{sec:main_results}.

%\subsection{Data, Training \& Evaluation Details}\label{sec:data_training_eval}

%\noindent \textbf{UMLS Data.} 
\paragraph{UMLS Data.} 
We rely on the UMLS (2020AA) as our \textsc{Sap} fine-tuning data, leveraging synonyms in all available languages. The full multilingual fine-tuning data comprises $\approx$15M biomedical entity names associated with $\approx$4.2M individual CUIs. As expected, English is dominant (69.6\% of all 15M names), followed by Spanish (10.7\%) and French (2.2\%). The full stats are in \Cref{sec:appendix_umls}. 

%\vspace{1.2mm}
%\noindent \textbf{Translation Data.}  
\paragraph{Translation Data.}  
We use (a) ``muse'' word translations \cite{lample2018word}, and (b) the parallel Wikipedia article titles (phrase-level translations; referred to as ``wt''). We also list results when using ``muse'' and ``wt'' combined (``wt+ muse''). 

%% (IV, Too verbose)
%% We investigated word-, phrase- and sentence-level bitexts and found that word-/phrase-level data generally works the better than sentence-level ones. This is rather intuitive as \xlbel evaluates on the granularity of word-/phrase-level.
%\vspace{1.2mm}
%\noindent \textbf{Training and Evaluation Details.}
\paragraph{Training and Evaluation Details.}
Our \textsc{Sap} fine-tuning largely follows \citet{liu2020self}; we refer to the original work and the Appendix for further technical details. The evaluation measure is standard Precision${@1}$ and Precision${@5}$. In all experiments, \textsc{Sap} always denotes fine-tuning of a base LM with UMLS data. \texttt{[CLS]} of the last layer's output is used as the final representation \cite{liu2020self}. Without explicit mentioning, we use the \textsc{Base} variants of all monolingual and multilingual LMs. At inference, given a query representation, a nearest neighbour search is used to rank all candidates' representations. We restrict the target ontology to only include CUIs that appear in WikiMed (62,531 CUIs, 399,931 entity names).

%% For, learning from bitext, we use an epoch of 5 and a batch size of 256 (128 bitext pairs). The other details are the same as pretraining.\footnote{More details are availible in \Cref{Table:search_space}.} 

%% (denoted as $@1$ and $@5$)

%% unless stated otherwise.\footnote{``\textsc{nospec}'' in the subscript denotes that the mean-pooling approach, recommended by \citet{vulic-etal-2020-probing}, is used.}.

\subsection{Main Results and Discussion}\label{sec:main_results}

%\stitle{Multilingual UMLS Knowledge Always Helps.}
\paragraph{Multilingual UMLS Knowledge Always Helps (\Cref{tab:model_sap}).}
\Cref{tab:model_sap} summarises the results of applying multilingual \textsc{Sap} fine-tuning based on UMLS knowledge on a wide variety of monolingual, multilingual, and in-domain pretrained encoders. Injecting UMLS knowledge is consistently beneficial to the models' performance on \xlbel across all languages and across all base encoders. Using multilingual UMLS synonyms to \textsc{Sap}-fine-tune the biomedical \textsc{PubMedBert} (\textsc{SapBert}$_{\text{all\_syn}}$) instead of English-only synonyms (\textsc{SapBert}) improves its performance across the board. \textsc{Sap}-ing monolingual \textsc{Bert}s for each language also yields substantial gains across all languages; the only exception is Thai (\tha), which is not represented in UMLS. Fine-tuning multilingual models \mbert and \xlmr leads to even larger relative gains. %on average.

%First, we tested \citet{liu2020self}'s approach of applying \textsc{Sap} on \textsc{PubMedBert} (\textsc{SapBert}). Instead of using only English synonyms as in the original work, we also created a variant that is pretrained on all synonyms in UMLS (\textsc{SapBert}$_{\text{all\_syn}}$). Performance gains is obtained upon almost all metrics of all languages.

%Secondly, we pick one language-specific pretrained \textsc{Bert} for every language and applied \textsc{Sap} with all synonyms to each of them. Very significant gains are observed across almost all languages (except for Thai, which has no synonyms existed in UMLS). Thirdly, we experiment with the most popular cross-lingual models \textsc{mBert} and \textsc{XLMR}. After applying \textsc{Sap}, even larger gains are observed on average. (We will discuss the variance across languages in following paragraphs.)

%\stitle{Large (cross-lingual) models are domain experts.} 

%\vspace{1.2mm}
%\noindent \textbf{Performance across Languages.} 
\paragraph{Performance across Languages (\Cref{tab:model_sap}).}
UMLS data is heavily biased towards Romance and Germanic languages. As a result, for languages more similar to these families, monolingual LMs (upper half, \Cref{tab:model_sap}) are on par or outperform multilingual LMs (lower half, \Cref{tab:model_sap}). However, for other (distant) languages (e.g., \ko, \zh, \ja, \tha), the opposite holds. For instance, on \tha, \textsc{XLMR+Sap}$_{\text{all\_syn}}$ outperforms \textsc{thBert+Sap}$_{\text{all\_syn}}$ by 20\% Precision${@1}$.

\begin{table}[!t]
\scriptsize
\setlength{\tabcolsep}{1.2pt}
%\vspace*{-0.3cm}
%\renewcommand{\arraystretch}{0.9}
%\vspace{-0.5em}
%\vspace{-0.5em}
\centering
\resizebox{0.99\columnwidth}{!}{
\begin{tabular}{llccccccccccccccccccccccccccccc}
\toprule
  language$\rightarrow$& \multicolumn{2}{c}{\es} & $\ $ &  \multicolumn{2}{c}{\de} &  $\ $ & \multicolumn{2}{c}{\ru} & $\ $ & 
  \multicolumn{2}{c}{\ko} & $\ $ & \multicolumn{2}{c}{\bf avg}  \\
 \cmidrule{2-3}\cmidrule{5-6} \cmidrule{8-9} \cmidrule{11-12} \cmidrule{14-15} \cmidrule{17-18} 
  model$\downarrow$ &  {\scriptsize @1} &\scriptsize @5 & &\scriptsize @1 &\scriptsize @5 & &\scriptsize @1 &\scriptsize @5 & &\scriptsize @1 &\scriptsize @5 & &\scriptsize @1 &\scriptsize @5 \\
  \midrule
\textsc{mBert} \\
 \rowcolor{cyan!10}
\textsc{ + Sap}$_{\text{en\_syn}}$ & 50.6 & 55.8 && 26.0 & 29.6 && 10.1 & 12.6 && 2.7 & 3.2 && 22.4 & 25.3 \\ 
 \rowcolor{cyan!10}
\textsc{ + Sap}$_{\text{\{\$LANG\}\_syn}}$ & 57.1 & 62.8 && 28.9 & 33.6 && 25.8 & 31.7 && 2.1 & 2.6 && 28.5 & 32.7\\
 \rowcolor{cyan!10}
\textsc{ + Sap}$_{\text{en+\{\$LANG\}\_syn}}$ & 61.1 & \textbf{68.5} && \textbf{35.2} & \textbf{39.8} &&  \textbf{35.6} & \textbf{40.9} && 14.4 & 16.3 && \textbf{36.6} & \textbf{41.4} \\
 \rowcolor{cyan!10}
\textsc{ + Sap}$_{\text{all\_syn}}$ & \textbf{61.4} & 67.0 && 33.4 & 37.8 && 35.1 & 40.3 && \textbf{15.1} & \textbf{17.6} && 36.6 & 40.7 \\
   \midrule
\textsc{XLMR} \\
 \rowcolor{cyan!10}
\textsc{ + Sap}$_{\text{en\_syn}}$ & 47.9 & 53.5 && 27.6 & 32.0 && 21.8 & 25.9 && 4.5 & 6.7 && 25.5 & 29.5 \\ 
 \rowcolor{cyan!10}
\textsc{ + Sap}$_{\text{\{\$LANG\}\_syn}}$ & 52.9 & 55.8 && 25.9 & 30.4 && 28.7 & 34.2 && 2.4 & 2.9 && 24.5 & 30.8 \\
 \rowcolor{cyan!10}
\textsc{ + Sap}$_{\text{en+\{\$LANG\}\_syn}}$ & 55.8 & 62.5 && 27.7 & 32.3 && \textbf{36.4} & \textbf{42.2} && 15.8 & 19.8 && 33.9 & 39.2 \\
 \rowcolor{cyan!10}
\textsc{ + Sap}$_{\text{all\_syn}}$ & \textbf{56.4} & \textbf{62.7} && \textbf{31.8} & \textbf{37.3} && 35.4 & 41.2 && \textbf{16.7} & \textbf{21.4} && \textbf{35.1} & \textbf{40.7}\\
\bottomrule
\end{tabular}
}%
%\vspace{-1.0mm}
\caption{Varying UMLS synonymy sets. }
%\vspace{-2.0mm}
\label{tab:the_more_the_better}
\end{table}

%\vspace{1.2mm}
%\noindent \textbf{General Translation Knowledge is Useful.}
\paragraph{General Translation Knowledge is Useful (\Cref{tab:bitext_helps}).}
\Cref{tab:bitext_helps} summarises the results where we continue training on general translation data (\S\ref{sec:sap_bitext}) after the previous UMLS-based \textsc{Sap}. With this variant, base multilingual LMs become powerful multilingual biomedical experts. We observe additional strong gains (cf., \Cref{tab:model_sap}) with out-of-domain translation data: e.g., for \mbert the gains range from 2.4\% to 12.7\% on all languages except \es. For \xlmr, we report Precision$@1$ boosts of $>$10\% on \ru, \tr, \ko, \tha with \textsc{XLMR+Sap}$_{\text{en\_syn}}$, and similar but smaller gains also with \textsc{XLMR+Sap}$_{\text{all\_syn}}$.

We stress the case of \tha, not covered in UMLS. Precision$@1$ rises from 11.5\% (\textsc{XLMR+Sap}$_{\text{en\_syn}}$) to 30.9\%$^{\uparrow19.4\%}$ (\textsc{XLMR+Sap}$_{\text{all\_syn}}$(+en-th wt+ muse)), achieved through the synergistic effect of both knowledge types: \textbf{1) UMLS synonyms in other languages} push the scores to 20.6\%$^{\uparrow9.1\%}$; \textbf{2) translation knowledge} increases it further to 30.9\%$^{\uparrow10.3\%}$. In general, these results suggest that both external in-domain knowledge and general-domain translations boost the performance in resource-poor languages. 

%\vspace{1.2mm}
\paragraph{The More the Better (\Cref{tab:the_more_the_better})?} According to \Cref{tab:the_more_the_better} (lower half), it holds almost universally that $\text{all\_syn} > \text{en+\{\$LANG\}\_syn} > \text{en\_syn}/\text{\{\$LANG\}\_syn}$ on \textsc{XLMR}, that is, it seems that more in-domain knowledge (even in non-related languages) benefit cross-lingual transfer. However, for \textsc{mBert} (\Cref{tab:the_more_the_better}, upper half), the trend is less clear, with $\text{en+\{\$LANG\}\_syn}$ sometimes outperforming the $\text{all\_syn}$ variant. Despite modest performance differences, this suggests that the choice of source languages for knowledge transfer also plays a role; this warrants further investigations in future work. 

%This suggests knowledge transfer is harder in many language pairs for \textsc{mBert}.

\paragraph{Are Large Models (Cross-Lingual) Domain Experts (\Cref{tab:large_xlmr})?}
We also investigate the \textsc{Large} variant of \textsc{XLMR}, and compare it to its \textsc{Base} variant. 
On English, \textsc{XLMR}$_{\textsc{Large}}$ gets 73.0\% Precision$@1$, being in the same range as \textsc{SapBert} (78.7\%), without \textsc{Sap}-tuning (\Cref{tab:large_xlmr}). The scores without \textsc{Sap} fine-tuning on \textsc{XLMR}$_{\textsc{Large}}$, although much higher than of its \textsc{Base} variant, decrease on other (`non-English') languages. At the same time, note that \textsc{XLMR Base} achieves random-level performance without \textsc{Sap}-tuning. After \textsc{Sap} fine-tuning, on average, \textsc{XLMR}$_{\textsc{Large}}$\textsc{+Sap} still outperforms \textsc{Base} models, but the gap is much smaller: e.g., we note that the performance of the two \textsc{Sap}-ed models is on par in English. This suggests that with sufficient knowledge injection, the underlying base model is less important (English); however, when the external data are scarce (other languages beyond English), a heavily parameterised large pretrained encoder can boost knowledge transfer to resource-poor languages.
%Most surprisingly, it even beats \textsc{SapBert} (first trained on both PubMed papers and then UMLS) on the English set of \xlbel-\textsc{hard}. This suggests that the \textsc{Large} language models are likely to contain even more domain knowledge than domain-specific models with less parameters. 

\begin{table}[!t]
\small
\setlength{\tabcolsep}{3.0pt}
%\vspace*{-0.3cm}
%\renewcommand{\arraystretch}{0.9}
\centering
%\resizebox{0.99\columnwidth}{!}{
\begin{tabular}{llccccccccccccccccccccccccccccc}
\toprule
  data split$\rightarrow$& \multicolumn{2}{c}{\en} &  &  \multicolumn{2}{c}{\bf avg} \\ % & &  \multicolumn{2}{c}{\textsc{hard} (\en)} & &  \multicolumn{2}{c}{\textsc{hard} ({\bf avg})}   \\
 \cmidrule{2-3}\cmidrule{5-6} % \cmidrule{8-9}\cmidrule{11-12} 
  model$\downarrow$ &  {\scriptsize @1} &\scriptsize @5 &&   {\scriptsize @1} &\scriptsize @5 \\ %&& {\scriptsize @1} &\scriptsize @5  && {\scriptsize @1} &\scriptsize @5 \\
  \midrule
\rowcolor{black!10}
  %\textsc{SapBert} & 78.7 & 81.6 && 19.9 & 21.6 && 52.7 & 59.1 && & \\
  %\hline
\rowcolor{black!10}
\textsc{XLMR} & 1.0 & 2.0 & & 0.2 & 0.5 \\ %&& 0.1 & 0.7 && 0.3 & 0.7 \\
   \rowcolor{cyan!10}
\textsc{ + Sap}$_{\text{all\_syn}}$ & 78.2 & 81.0 && 34.3 & 39.3 \\ %&& 50.6 & 59.2 && 34.1 & 40.4  \\
\hline
   \rowcolor{black!10}
\textsc{XLMR}$_{\textsc{Large}}$ & 73.0 & 75.0 &&12.3 & 13.3 \\ % && 38.3 & 42.6 && 7.0 & 8.2 \\
   \rowcolor{cyan!10}
\textsc{ + Sap}$_{\text{all\_syn}}$ & \textbf{78.3} & \textbf{81.3} && \textbf{39.0} & \textbf{44.2} \\ %&& 51.2 & 58.2 && 39.5 & 45.9 \\
\bottomrule
\end{tabular}
%}%
%\vspace{-1.0mm}
\caption{Comparing \textsc{Base} and \textsc{Large} models on \xlbel. Both \en results and {\bf avg} across all languages are reported. Full table available in Appendix \Cref{tab:large_models}.}
%\vspace{-2.0mm}
\label{tab:large_xlmr}
\end{table}

\section{Conclusion}
We have introduced a novel cross-lingual biomedical entity task (\xlbel), establishing a wide-coverage and reliable evaluation benchmark for cross-lingual entity representations in the biomedical domain in 10 languages, and have evaluated current SotA biomedical entity representations on \xlbel. We have also presented an effective transfer learning scheme that leverages general-domain translations to improve the cross-lingual ability of domain-specialised representation models. We hope that our work will inspire more research on multilingual \textit{and} domain-specialised representation learning in the future.

%Our results show as much as 20\% of Precision$_{@1}$ enhancement on \xlbel, without leveraging any in-domain parallel data.

%% After a thorough investigation into knowledge-agnostic/enhanced and mono-/multilingual pretrained LMs, we found \xlbel to be very hard, especially for languages distant to English. 

\section*{Acknowledgements}
We thank the three reviewers and the AC for their insightful comments and suggestions. FL is supported by Grace \& Thomas C.H. Chan Cambridge Scholarship. IV and AK are supported by the ERC Consolidator Grant LEXICAL (no. 648909) awarded to AK. NC kindly acknowledges grant-in-aid support from the UK ESRC for project EPI-AI (ES/T012277/1). 

\bibliographystyle{acl_natbib}
\bibliography{anthology,acl2021}

\begin{thebibliography}{27}
\expandafter\ifx\csname natexlab\endcsname\relax\def\natexlab#1{#1}\fi

\bibitem[{Alsentzer et~al.(2019)Alsentzer, Murphy, Boag, Weng, Jindi, Naumann,
  and McDermott}]{clinicalbert}
Emily Alsentzer, John Murphy, William Boag, Wei-Hung Weng, Di~Jindi, Tristan
  Naumann, and Matthew McDermott. 2019.
\newblock \href {https://doi.org/10.18653/v1/W19-1909} {Publicly available
  clinical {BERT} embeddings}.
\newblock In \emph{Proceedings of the 2nd Clinical Natural Language Processing
  Workshop}, pages 72--78, Minneapolis, Minnesota, USA. Association for
  Computational Linguistics.

\bibitem[{Artetxe and Schwenk(2019)}]{artetxe2019massively}
Mikel Artetxe and Holger Schwenk. 2019.
\newblock \href {https://doi.org/10.1162/tacl_a_00288} {Massively multilingual
  sentence embeddings for zero-shot cross-lingual transfer and beyond}.
\newblock \emph{Transactions of the Association for Computational Linguistics},
  7:597--610.

\bibitem[{Basaldella et~al.(2020)Basaldella, Liu, Shareghi, and
  Collier}]{cometa}
Marco Basaldella, Fangyu Liu, Ehsan Shareghi, and Nigel Collier. 2020.
\newblock \href {https://doi.org/10.18653/v1/2020.emnlp-main.253} {{COMETA}: A
  corpus for medical entity linking in the social media}.
\newblock In \emph{Proceedings of the 2020 Conference on Empirical Methods in
  Natural Language Processing (EMNLP)}, pages 3122--3137, Online. Association
  for Computational Linguistics.

\bibitem[{Bawden et~al.(2019)Bawden, Bretonnel~Cohen, Grozea, Jimeno~Yepes,
  Kittner, Krallinger, Mah, Neveol, Neves, Soares, Siu, Verspoor, and
  Vicente~Navarro}]{bawden-etal-2019-findings}
Rachel Bawden, Kevin Bretonnel~Cohen, Cristian Grozea, Antonio Jimeno~Yepes,
  Madeleine Kittner, Martin Krallinger, Nancy Mah, Aurelie Neveol, Mariana
  Neves, Felipe Soares, Amy Siu, Karin Verspoor, and Maika Vicente~Navarro.
  2019.
\newblock \href {https://doi.org/10.18653/v1/W19-5403} {Findings of the {WMT}
  2019 biomedical translation shared task: Evaluation for {MEDLINE} abstracts
  and biomedical terminologies}.
\newblock In \emph{Proceedings of the Fourth Conference on Machine Translation
  (Volume 3: Shared Task Papers, Day 2)}, pages 29--53, Florence, Italy.
  Association for Computational Linguistics.

\bibitem[{Bodenreider(2004)}]{bodenreider2004unified}
Olivier Bodenreider. 2004.
\newblock \href
  {https://www.ncbi.nlm.nih.gov/pmc/articles/PMC308795/pdf/gkh061.pdf} {The
  unified medical language system ({UMLS}): integrating biomedical
  terminology}.
\newblock \emph{Nucleic Acids Research}, 32:D267--D270.

\bibitem[{Conneau et~al.(2020)Conneau, Khandelwal, Goyal, Chaudhary, Wenzek,
  Guzm{\'a}n, Grave, Ott, Zettlemoyer, and Stoyanov}]{conneau2020unsupervised}
Alexis Conneau, Kartikay Khandelwal, Naman Goyal, Vishrav Chaudhary, Guillaume
  Wenzek, Francisco Guzm{\'a}n, Edouard Grave, Myle Ott, Luke Zettlemoyer, and
  Veselin Stoyanov. 2020.
\newblock \href {https://doi.org/10.18653/v1/2020.acl-main.747} {Unsupervised
  cross-lingual representation learning at scale}.
\newblock In \emph{Proceedings of the 58th Annual Meeting of the Association
  for Computational Linguistics}, pages 8440--8451, Online. Association for
  Computational Linguistics.

\bibitem[{Devlin et~al.(2019)Devlin, Chang, Lee, and
  Toutanova}]{devlin2019bert}
Jacob Devlin, Ming-Wei Chang, Kenton Lee, and Kristina Toutanova. 2019.
\newblock \href {https://doi.org/10.18653/v1/N19-1423} {{BERT}: Pre-training of
  deep bidirectional transformers for language understanding}.
\newblock In \emph{Proceedings of the 2019 Conference of the North {A}merican
  Chapter of the Association for Computational Linguistics: Human Language
  Technologies, Volume 1 (Long and Short Papers)}, pages 4171--4186,
  Minneapolis, Minnesota. Association for Computational Linguistics.

\bibitem[{Gu et~al.(2020)Gu, Tinn, Cheng, Lucas, Usuyama, Liu, Naumann, Gao,
  and Poon}]{pubmedbert}
Yu~Gu, Robert Tinn, Hao Cheng, Michael Lucas, Naoto Usuyama, Xiaodong Liu,
  Tristan Naumann, Jianfeng Gao, and Hoifung Poon. 2020.
\newblock \href {https://arxiv.org/pdf/2007.15779.pdf} {Domain-specific
  language model pretraining for biomedical natural language processing}.
\newblock \emph{arXiv:2007.15779}.

\bibitem[{Hu et~al.(2020)Hu, Ruder, Siddhant, Neubig, Firat, and
  Johnson}]{Hu:2020icml}
Junjie Hu, Sebastian Ruder, Aditya Siddhant, Graham Neubig, Orhan Firat, and
  Melvin Johnson. 2020.
\newblock \href {http://proceedings.mlr.press/v119/hu20b.html} {{XTREME:} {A}
  massively multilingual multi-task benchmark for evaluating cross-lingual
  generalisation}.
\newblock In \emph{Proceedings of the 37th International Conference on Machine
  Learning, {ICML} 2020, 13-18 July 2020, Virtual Event}, volume 119 of
  \emph{Proceedings of Machine Learning Research}, pages 4411--4421. {PMLR}.

\bibitem[{Lample et~al.(2018)Lample, Conneau, Ranzato, Denoyer, and
  J{\'{e}}gou}]{lample2018word}
Guillaume Lample, Alexis Conneau, Marc'Aurelio Ranzato, Ludovic Denoyer, and
  Herv{\'{e}} J{\'{e}}gou. 2018.
\newblock \href {https://openreview.net/forum?id=H196sainb} {Word translation
  without parallel data}.
\newblock In \emph{6th International Conference on Learning Representations,
  {ICLR} 2018, Vancouver, BC, Canada, April 30 - May 3, 2018, Conference Track
  Proceedings}. OpenReview.net.

\bibitem[{Lauscher et~al.(2020)Lauscher, Vuli{\'c}, Ponti, Korhonen, and
  Glava{\v{s}}}]{Lauscher:2020coling}
Anne Lauscher, Ivan Vuli{\'c}, Edoardo~Maria Ponti, Anna Korhonen, and Goran
  Glava{\v{s}}. 2020.
\newblock \href {https://doi.org/10.18653/v1/2020.coling-main.118}
  {Specializing unsupervised pretraining models for word-level semantic
  similarity}.
\newblock In \emph{Proceedings of the 28th International Conference on
  Computational Linguistics}, pages 1371--1383, Barcelona, Spain (Online).
  International Committee on Computational Linguistics.

\bibitem[{Lee et~al.(2020)Lee, Yoon, Kim, Kim, Kim, So, and
  Kang}]{lee2020biobert}
Jinhyuk Lee, Wonjin Yoon, Sungdong Kim, Donghyeon Kim, Sunkyu Kim, Chan~Ho So,
  and Jaewoo Kang. 2020.
\newblock \href
  {https://academic.oup.com/bioinformatics/article/36/4/1234/5566506}
  {{BioBERT}: a pre-trained biomedical language representation model for
  biomedical text mining}.
\newblock \emph{Bioinformatics}, 36(4):1234--1240.

\bibitem[{Lee et~al.(2016)Lee, Kim, Lee, Choi, Kim, Jeon, Lim, Choi, Kim, Tan
  et~al.}]{lee2016best}
Sunwon Lee, Donghyeon Kim, Kyubum Lee, Jaehoon Choi, Seongsoon Kim, Minji Jeon,
  Sangrak Lim, Donghee Choi, Sunkyu Kim, Aik-Choon Tan, et~al. 2016.
\newblock \href
  {https://journals.plos.org/plosone/article?id=10.1371/journal.pone.0164680}
  {{BEST}: next-generation biomedical entity search tool for knowledge
  discovery from biomedical literature}.
\newblock \emph{PloS one}, 11(10):e0164680.

\bibitem[{Levine et~al.(2020)Levine, Lenz, Dagan, Ram, Padnos, Sharir,
  Shalev-Shwartz, Shashua, and Shoham}]{Levine:2020acl}
Yoav Levine, Barak Lenz, Or~Dagan, Ori Ram, Dan Padnos, Or~Sharir, Shai
  Shalev-Shwartz, Amnon Shashua, and Yoav Shoham. 2020.
\newblock \href {https://doi.org/10.18653/v1/2020.acl-main.423} {{S}ense{BERT}:
  Driving some sense into {BERT}}.
\newblock In \emph{Proceedings of the 58th Annual Meeting of the Association
  for Computational Linguistics}, pages 4656--4667, Online. Association for
  Computational Linguistics.

\bibitem[{Liu et~al.(2021)Liu, Shareghi, Meng, Basaldella, and
  Collier}]{liu2020self}
Fangyu Liu, Ehsan Shareghi, Zaiqiao Meng, Marco Basaldella, and Nigel Collier.
  2021.
\newblock \href {https://www.aclweb.org/anthology/2021.naacl-main.334}
  {Self-alignment pretraining for biomedical entity representations}.
\newblock In \emph{Proceedings of the 2021 Conference of the North American
  Chapter of the Association for Computational Linguistics: Human Language
  Technologies}, pages 4228--4238, Online. Association for Computational
  Linguistics.

\bibitem[{Liu et~al.(2019)Liu, Ott, Goyal, Du, Joshi, Chen, Levy, Lewis,
  Zettlemoyer, and Stoyanov}]{liu2019roberta}
Yinhan Liu, Myle Ott, Naman Goyal, Jingfei Du, Mandar Joshi, Danqi Chen, Omer
  Levy, Mike Lewis, Luke Zettlemoyer, and Veselin Stoyanov. 2019.
\newblock \href {https://arxiv.org/pdf/1907.11692.pdf} {Roberta: A robustly
  optimized bert pretraining approach}.
\newblock \emph{arXiv preprint arXiv:1907.11692}.

\bibitem[{McNamee et~al.(2011)McNamee, Mayfield, Lawrie, Oard, and
  Doermann}]{mcnamee2011cross}
Paul McNamee, James Mayfield, Dawn Lawrie, Douglas Oard, and David Doermann.
  2011.
\newblock \href {https://www.aclweb.org/anthology/I11-1029} {Cross-language
  entity linking}.
\newblock In \emph{Proceedings of 5th International Joint Conference on Natural
  Language Processing}, pages 255--263, Chiang Mai, Thailand. Asian Federation
  of Natural Language Processing.

\bibitem[{Peng et~al.(2019)Peng, Yan, and Lu}]{peng2019transfer}
Yifan Peng, Shankai Yan, and Zhiyong Lu. 2019.
\newblock \href {https://doi.org/10.18653/v1/W19-5006} {Transfer learning in
  biomedical natural language processing: An evaluation of {BERT} and {ELM}o on
  ten benchmarking datasets}.
\newblock In \emph{Proceedings of the 18th BioNLP Workshop and Shared Task},
  pages 58--65, Florence, Italy. Association for Computational Linguistics.

\bibitem[{Reimers and Gurevych(2020)}]{reimers2020making}
Nils Reimers and Iryna Gurevych. 2020.
\newblock \href {https://doi.org/10.18653/v1/2020.emnlp-main.365} {Making
  monolingual sentence embeddings multilingual using knowledge distillation}.
\newblock In \emph{Proceedings of the 2020 Conference on Empirical Methods in
  Natural Language Processing (EMNLP)}, pages 4512--4525, Online. Association
  for Computational Linguistics.

\bibitem[{Sung et~al.(2020)Sung, Jeon, Lee, and
  Kang}]{sung-etal-2020-biomedical}
Mujeen Sung, Hwisang Jeon, Jinhyuk Lee, and Jaewoo Kang. 2020.
\newblock \href {https://doi.org/10.18653/v1/2020.acl-main.335} {Biomedical
  entity representations with synonym marginalization}.
\newblock In \emph{Proceedings of the 58th Annual Meeting of the Association
  for Computational Linguistics}, pages 3641--3650, Online. Association for
  Computational Linguistics.

\bibitem[{Tsai and Roth(2016)}]{tsai2016cross}
Chen-Tse Tsai and Dan Roth. 2016.
\newblock \href {https://doi.org/10.18653/v1/N16-1072} {Cross-lingual
  wikification using multilingual embeddings}.
\newblock In \emph{Proceedings of the 2016 Conference of the North {A}merican
  Chapter of the Association for Computational Linguistics: Human Language
  Technologies}, pages 589--598, San Diego, California. Association for
  Computational Linguistics.

\bibitem[{Tutubalina et~al.(2020)Tutubalina, Kadurin, and
  Miftahutdinov}]{tutubalina-etal-2020-fair}
Elena Tutubalina, Artur Kadurin, and Zulfat Miftahutdinov. 2020.
\newblock \href {https://www.aclweb.org/anthology/2020.coling-main.588} {Fair
  evaluation in concept normalization: a large-scale comparative analysis for
  {BERT}-based models}.
\newblock In \emph{Proceedings of the 28th International Conference on
  Computational Linguistics}, pages 6710--6716, Barcelona, Spain (Online).
  International Committee on Computational Linguistics.

\bibitem[{{Vashishth} et~al.(2020){Vashishth}, {Joshi}, {Newman-Griffis},
  {Dutt}, and {Rose}}]{medtype2020}
Shikhar {Vashishth}, Rishabh {Joshi}, Denis {Newman-Griffis}, Ritam {Dutt}, and
  Carolyn {Rose}. 2020.
\newblock \href {http://arxiv.org/abs/2005.00460} {{MedType: Improving Medical
  Entity Linking with Semantic Type Prediction}}.
\newblock \emph{arXiv e-prints}, page arXiv:2005.00460.

\bibitem[{Vaswani et~al.(2017)Vaswani, Shazeer, Parmar, Uszkoreit, Jones,
  Gomez, Kaiser, and Polosukhin}]{Vaswani:2017nips}
Ashish Vaswani, Noam Shazeer, Niki Parmar, Jakob Uszkoreit, Llion Jones,
  Aidan~N. Gomez, Lukasz Kaiser, and Illia Polosukhin. 2017.
\newblock \href
  {https://proceedings.neurips.cc/paper/2017/hash/3f5ee243547dee91fbd053c1c4a845aa-Abstract.html}
  {Attention is all you need}.
\newblock In \emph{Advances in Neural Information Processing Systems 30: Annual
  Conference on Neural Information Processing Systems 2017, December 4-9, 2017,
  Long Beach, CA, {USA}}, pages 5998--6008.

\bibitem[{Wang et~al.(2019)Wang, Han, Huang, Dong, and Scott}]{wang2019multi}
Xun Wang, Xintong Han, Weilin Huang, Dengke Dong, and Matthew~R. Scott. 2019.
\newblock \href {https://doi.org/10.1109/CVPR.2019.00516} {Multi-similarity
  loss with general pair weighting for deep metric learning}.
\newblock In \emph{{IEEE} Conference on Computer Vision and Pattern
  Recognition, {CVPR} 2019, Long Beach, CA, USA, June 16-20, 2019}, pages
  5022--5030. Computer Vision Foundation / {IEEE}.

\bibitem[{Wang et~al.(2018)Wang, Liu, Afzal, Rastegar-Mojarad, Wang, Shen,
  Kingsbury, and Liu}]{wang2018comparison}
Yanshan Wang, Sijia Liu, Naveed Afzal, Majid Rastegar-Mojarad, Liwei Wang,
  Feichen Shen, Paul Kingsbury, and Hongfang Liu. 2018.
\newblock \href
  {https://www.sciencedirect.com/science/article/pii/S1532046418301825} {A
  comparison of word embeddings for the biomedical natural language
  processing}.
\newblock \emph{Journal of biomedical informatics}, 87:12--20.

\bibitem[{Zhang et~al.(2019)Zhang, Han, Liu, Jiang, Sun, and
  Liu}]{Zhang:2019acl}
Zhengyan Zhang, Xu~Han, Zhiyuan Liu, Xin Jiang, Maosong Sun, and Qun Liu. 2019.
\newblock \href {https://doi.org/10.18653/v1/P19-1139} {{ERNIE}: Enhanced
  language representation with informative entities}.
\newblock In \emph{Proceedings of the 57th Annual Meeting of the Association
  for Computational Linguistics}, pages 1441--1451, Florence, Italy.
  Association for Computational Linguistics.

\end{thebibliography}

\clearpage

\appendix
\section{Appendix A}\label{sec:app_a}

% \begin{table}[!h]
%     \centering
%     \begin{tabular}{c|c}
%     \toprule
%          en & English  \\
%          es & Spanish \\
%          de & German \\
%          fi & Finnish \\
%          %eu & Basque \\
%          ru & Russian \\
%          tr & Turkish \\
%          ko & Korean \\
%          zh & Chinese \\
%          ja & Japanese \\
%          th & Thai \\
%     \bottomrule
%     \end{tabular}
%     \caption{Language abbreviations.}
%   % \label{tab:my_label}
% \end{table}

\subsection{\xlbel: Full Statistics}\label{sec:appendix_xlbel}
\Cref{tab:xlbel} in the main paper summarises the key statistics of the \xlbel benchmark. It was extracted from the \texttt{20200601} version of Wikipedia dump. ``sentences'' refers to the number of sentences that contain biomedical mentions in the Wiki dump. ``unique titles (Wiki page)'' denotes the number of unique Wikipedia articles the biomedical mentions link to. ``mentions'' denotes the number of all biomedical mentions in the Wikipedia dump. ``unique mentions'' refers to the number of mentions after filtering out examples containing duplicated mention surface forms. ``unique mentions$_{\text{mention!=title}}$'' denotes the number of unique mentions that have surface forms different from the Wikipedia articles they link to. 
The 1k test sets for each language are then randomly selected from the examples in ``unique mentions$_{\text{mention!=title}}$''.

%The test sets are uploaded as a separate file. We will release the full dataset upon acceptance.

\iffalse
\begin{table*}[!t]
\small
\setlength{\tabcolsep}{1.8pt}
%\renewcommand{\arraystretch}{0.9}
\centering
\begin{tabular}{lcccccccccccc}
\toprule
 %\multirow{2}{*}{model} &  \multicolumn{3}{c}{FR$\rightarrow$EN}  \\
 %\cmidrule{2-4}
\#$\downarrow$, language$\rightarrow$ & \en & \es & \de & \fin & \ru & \tr & \ko & \zh & \ja & \tha\\
\midrule
sentences & - & 223,506 & 350,193 & 77,736 & 206,060 & 29,473 & 47,702 & 136,054 & 157,670 & 19,066\\
unique titles (Wiki page) & 60,598 & 37,935 & 24,059 & 15,182 & 21,044 & 5,251 & 10,618 & 17,972 & 11,002 & 4,541 \\ 
mentions & 1,067,083 & 204,253 & 431,781 & 105,182 & 221,383 & 29,958 & 60,979 & 197,317 & 220,452 & 31,177 \\
unique mentions & 121,669 & 25,169 & 44,390 & 26,184 & 28,302 & 4,110 & 9,032 & 24,825 & 21,949 & 5,064 \\
unique mentions$_{\text{mention!=title}}$ & 237,851 & 22,162 & 43,753 &  19,409 & 23,935 & 2,833 & 3,740 & 12,046 & 12,571 & 2,480\\
\bottomrule
\end{tabular}
%\vspace{-1.4em}
\caption{Construction of the \xlbel benchmark; key statistics. }
\label{tab:xlbel}
\end{table*}
\fi

\subsection{\xlbel: Selection of Languages}\label{sec:appendix_choice_of_lang}
Our goal is to select a diverse and representative sample of languages for the resource and evaluation from the full set of possibly supported languages. For this reason, we exclude some Romance and Germanic languages which were too similar to some languages already included in the resource (e.g., since we include Spanish as a representative of the Romance language, evaluating on related languages such as Portuguese or Italian would not yield additional and new insights, while it would just imply running additional experiments). The language list covers languages that are close to English (Spanish, German); languages that are very distant from English (Thai, Chinese, etc.); and also languages that are \textit{in the middle} (e.g., Turkish, which is typologically different, but shares a similar writing script with English).

The availability of biomedical texts in Wikipedia also slightly impacted our choice of languages. The overlapping entities of Wikipedia and UMLS are not evenly distributed in the biomedical domain. For example, since animal species are comprehensively encoded in UMLS, they become rather dominant for certain low-resource languages. We manually inspected the distribution of the covered entities in each language to ensure that they are indeed representative biomedical concepts. Languages with heavily skewed entity distributions are filtered out. E.g., biomedical concepts in Basque Wikipedia are heavily skewed towards plant and animal species (which are valid UMLS concepts but not representative enough). As a result, we dropped Basque as our evaluation language. The current 10 languages all have a reasonably fair distribution over biomedical concepts categories.

\subsection{UMLS Data Preparation}\label{sec:appendix_umls}
All our UMLS fine-tuning data for \textsc{Sap} is extracted from the \texttt{MRCONSO.RRF} file downloaded at \url{https://www.nlm.nih.gov/research/umls/licensedcontent/umlsarchives04.html#2020AA}. The extracted data includes 147,706,62 synonyms distributed in more than 20 languages. The detailed statistics are available in \Cref{tab:umls}.

\begin{table}[!ht]
    \centering
    \small
    \begin{tabular}{llll}
    \toprule
    code & language & \# synonyms & percentage \\
    \midrule
    \en & English & 10,277,246 & 69.6\% \\
    \es & Spanish & 1,575,109 & 10.7\% \\
    \ja & Japanese & 329,333 & 2.2\% \\
    \ru & Russian & 291,554 &  2.0\% \\
    \de & German & 231,098 & 1.6\% \\
    \ko & Korean & 145,865 & 1.0\% \\
    \zh & Chinese & 80,602 & 0.5\% \\
    \tr & Turkish & 51,328 & 0.3\% \\
    \fin & Finnish & 24,767 & 0.2\% \\
    \tha & Thai & 0 & 0.0\% \\
    \hdashline
    \fr & French & 428,406 & 2.9\% \\
    \pt & Portuguese & 309,448 & 2.1\% \\
    \nl & Dutch & 290,415 & 2.0\% \\
    \ita & Italian & 242,133 & 1.3\% \\ 
    \cs & Czech & 196,760 & 0.7\% \\
    \no & Norwegian & 63,075 & 0.4\%\\
    \pl & Polish & 51,778 & 0.4\% \\
    \et & Estonian & 31,107 & 0.2\% \\
    \sv & Swedish & 29,716 & 0.2\% \\
    \hr & Croatian & 10,035 & 0.1\% \\
    \el & Greek & 2,281 & $<$0.1\%  \\
    \lv & Latvian	& 1,405 & $<$0.1\% \\
    \midrule
    & Total & 147,706,62 & 100\% \\
    \bottomrule
    \end{tabular}
    \caption{The amount of UMLS synonyms per language. The first 10 languages are included in our \xlbel test languages. However, note that Thai has no UMLS data.}
    \label{tab:umls}
\end{table}

\subsection{Translation Data}\label{sec:appendix_bitext}
The full statistics of the used word and phrase translation data are listed in \Cref{tab:bitext_stat}. The ``muse'' word translations are downloaded from \url{https://github.com/facebookresearch/MUSE} while the Wikititle pairs (``wt'') are extracted by us, and are made publicly available.

\begin{table*}[!ht]
\small
\centering
\begin{tabular}{lcccccccccccc}
\toprule
\#$\downarrow$, language$\rightarrow$ & \en-\es & \en-\de & \en-\fin & \en-\ru & \en-\tr & \en-\ko & \en-\zh & \en-\ja & \en-\tha \\
\midrule
muse & 112,583 & 101,931 &  43,102 & 48,714 & 68,611 & 20,549 & 39,334 & 25,969 & 25,332 \\
wt & 1,079,547 & 1,241,104 & 338,284 & 886,760 & 260,392 & 319,492 & 638,900 & 547,923 & 107,398 \\
\bottomrule
\end{tabular}
%\vspace{-1.4em}
\caption{Statistics of muse word translations (``muse'') and Wikipedia title pairs (``wt'').}
\label{tab:bitext_stat}
\end{table*}

\subsection{Pretrained Encoders}\label{sec:appendix_huggingface_urls}
A complete listing of URLs for all used pretrained encoders hosted on \url{huggingface.co} is provided in \Cref{tab:model_url}. For monolingual models of each language, we made the best effort to select the most popular one (based on download counts).

\begin{table*}[!ht] %[!htbp]
\small
\setlength{\tabcolsep}{1pt}
%\vspace*{-0.3cm}
%\renewcommand{\arraystretch}{0.8}
\centering
\begin{tabular}{ll}
\toprule
model & URL \\
\midrule
 \multicolumn{2}{l}{\xspace\xspace\emph{monolingual models}} \\
 \midrule
\textsc{SapBert} & \url{https://huggingface.co/cambridgeltl/SapBERT-from-PubMedBERT-fulltext} \\
\textsc{esBert}  & \url{https://huggingface.co/dccuchile/bert-base-spanish-wwm-uncased} \\
\textsc{deBert}  &  \url{https://huggingface.co/dbmdz/bert-base-german-uncased} \\
\textsc{fiBert}  &  \url{https://huggingface.co/TurkuNLP/bert-base-finnish-uncased-v1} \\
\textsc{ruBert}  &  \url{https://huggingface.co/DeepPavlov/rubert-base-cased} \\
\textsc{trBert}  &  \url{https://huggingface.co/loodos/bert-base-turkish-uncased} \\
\textsc{krBert}  &  \url{https://huggingface.co/snunlp/KR-BERT-char16424} \\
\textsc{zhBert}  &  \url{https://huggingface.co/bert-base-chinese} \\
\textsc{jaBert}  &  \url{https://huggingface.co/cl-tohoku/bert-base-japanese} \\
\textsc{thBert}  &  \url{https://huggingface.co/monsoon-nlp/bert-base-thai} \\
\midrule
 \multicolumn{2}{l}{\xspace\xspace\emph{cross-lingual models}} \\
 \midrule
\textsc{mBert} & \url{https://huggingface.co/bert-base-multilingual-uncased} \\
\textsc{XLMR} & \url{https://huggingface.co/xlm-roberta-base} \\
\textsc{XLMR}$_{\textsc{Large}}$ & \url{https://huggingface.co/xlm-roberta-large} \\
\textsc{XLMR}$_{\textsc{Large-xnli}}$ & \url{https://huggingface.co/joeddav/xlm-roberta-large-xnli}\\
\textsc{XLMR}$_{\textsc{Large-squad2}}$ & \url{https://huggingface.co/deepset/xlm-roberta-large-squad2}\\
\bottomrule
\end{tabular}
\caption{A listing of HuggingFace URLs of all pretrained models used in this work.}
\label{tab:model_url}
\end{table*}

\subsection{Full Table for Comparing with \textsc{Large} Models}
\Cref{tab:large_models} list results across all languages for comparing \textsc{Base} and \textsc{Large} models.

\subsection{Future Work}\label{sec:future_work}

\stitle{Investigating Other Cross-Lingual Transfer Learning Schemes.} We also explored adapting multilingual sentence representation transfer techniques like \citet{reimers2020making} that leverage parallel data. However, we observed no improvement comparing to the main transfer scheme reported in the paper. We plan to investigate existing techniques more comprehensively, and benchmark more results on \xlbel in the future.

\begin{table*}[!ht]
\scriptsize
\setlength{\tabcolsep}{1.5pt}
%\vspace*{-0.3cm}
%\renewcommand{\arraystretch}{0.9}
%\vspace{-0.5em}
%\vspace{-0.5em}
\centering
\begin{tabular}{llccccccccccccccccccccccccccccccccc}
\toprule
  language$\rightarrow$& \multicolumn{2}{c}{\en} & $\ $ & \multicolumn{2}{c}{\es} & $\ $ & \multicolumn{2}{c}{\de} & $\ $ &  \multicolumn{2}{c}{\fin} & $\ $ &  \multicolumn{2}{c}{\ru} & $\ $ & 
  \multicolumn{2}{c}{\tr} & $\ $ &
  \multicolumn{2}{c}{\ko} & $\ $ & \multicolumn{2}{c}{\zh} & $\ $ &  \multicolumn{2}{c}{\ja} &  $\ $ &  \multicolumn{2}{c}{\tha} & $\ $ &  \multicolumn{2}{c}{\bf avg}\\
 \cmidrule{2-3}\cmidrule{5-6} \cmidrule{8-9} \cmidrule{11-12} \cmidrule{14-15} \cmidrule{17-18}\cmidrule{20-21}\cmidrule{23-24}\cmidrule{26-27}\cmidrule{29-30}\cmidrule{32-33}
  model$\downarrow$ &  {\scriptsize @1} &\scriptsize @5 & &\scriptsize @1 &\scriptsize @5 & &\scriptsize @1 &\scriptsize @5 & &\scriptsize @1 &\scriptsize @5 & &\scriptsize @1 &\scriptsize @5 & &\scriptsize @1 &\scriptsize @5 & &\scriptsize @1 &\scriptsize @5 & &\scriptsize @1 &\scriptsize @5 & &\scriptsize @1 &\scriptsize @5 & &\scriptsize @1 &\scriptsize @5  & &\scriptsize @1 &\scriptsize @5  \\
  \midrule
  \textsc{SapBert} & \textbf{78.7} & \textbf{81.6} && 47.3 & 51.4 && 22.7 & 24.7 && 8.2 & 10.2 && 5.8 & 6.0 && 26.4 & 29.7 && 2.0 & 2.4 && 1.9 & 2.2 && 3.0 & 3.2 && 3.1 & 3.4 && 19.9 & 21.6 \\
    \textsc{SapBert}$_{\text{all\_syn}}$ & 78.3 & 80.7 && 55.6 & 61.3 && 30.0 & 34.2 && 11.8 & 14.8 && 9.3 & 11.3 && 35.5 & 39.5 && 2.0 & 2.4 && 6.4 & 8.2 && 6.9 & 8.3 && 3.0 & 3.3 && 23.9 & 26.4 \\
%\hdashline
%\textsc{mBert}$_{\textsc{nospec}}$ & 35.8  & 49.4 && 15.0 & 24.1 && 7.1 & 10.4 && 1.7 & 2.6 && 2.3 & 3.6 && 10.8 & 15.3 &&  1.3 & 1.7 && 0.9 & 1.4 && 1.7 & 2.2 && 1.1 & 1.7 && 7.8 & 11.2 \\
  % \textsc{mBert-Large}$_{\textsc{nospec}}$ & & &&\\
\textsc{XLMR} & 1.0 & 2.0 && 0.3 & 0.7 && 0.0 & 0.1 && 0.1 & 0.2 && 0.1 & 0.2 && 0.4 & 0.5 && 0.0 & 0.3 && 0.1 & 0.2 && 0.2 & 0.4 && 0.0 & 0.1 && 0.2 & 0.5 \\
%\textsc{XLMR}$_{\textsc{nospec}}$ & 0.2 & 0.4 && 0.0 & 0.0 && 0.2 & 0.3 && 0.1 & 0.2 && 0.2 & 0.2 && 0.0 & 0.0 && 0.0 & 0.2 && 0.1 & 0.1 && 0.0 & 0.2 && 0.1 & 0.3 && 0.1 & 0.2 \\
\textsc{XLMR + Sap}$_{\text{all\_syn}}$ & 78.2 & 81.0 && 56.4 & 62.7 && 31.8 & 37.3 && 18.6 & 22.2 && 35.4 & 41.2 && 42.8 & 48.9 && 16.7 & 21.4 && 18.8 & 23.0 && 24.0 & 28.1 && 20.6 & 27.5 && 34.3 & 39.3 \\
\midrule
\textsc{XLMR}$_{\textsc{Large}}$  & 73.0 & 75.0 && 20.7 & 24.6 && 7.8 & 9.1 && 1.9 & 2.7 && 3.0 & 3.3 && 11.8 & 13.5 && 1.2 & 1.2 && 0.7 & 0.9 && 1.6 & 1.8 && 0.9 & 1.2 && 12.3 & 13.3 \\
%\textsc{XLMR-Large}$_{\textsc{nospec}}$ & 27.0 &  29.8 && 10.5 & 13.1 && 3.7 & 4.4 && 1.3 & 1.6 \\
%\hdashline
\textsc{XLMR}$_{\textsc{Large-xnli}}$ & 72.6 & 75.1 && 30.1 & 33.5 && 10.7 & 12.2 && 3.4 & 4.6 && 5.9 & 7.4 && 16.4 & 18.4 && 1.9 & 2.6 && 1.3 & 2.0 && 2.0 & 2.5 && 1.3 & 2.0 && 14.6 & 16.0 \\
\textsc{XLMR}$_{\textsc{Large-squad2}}$ & 74.6 & 76.2 && 31.4 & 35.3 && 11.9 & 13.2 && 3.5 & 4.4 && 5.2 & 6.5 && 16.9 & 19.2 && 1.4 & 1.5 && 0.6 & 0.9 && 1.8 & 2.1 && 2.0 & 2.3 && 14.9 & 16.2 \\
\textsc{XLMR}$_{\textsc{Large}}$ + \textsc{Sap}$_{\text{all\_syn}}$ & 78.3 & 81.3 && \textbf{61.0} & \textbf{66.8} && \textbf{35.0} & \textbf{40.0} && \textbf{25.2} & \textbf{29.2} && \textbf{41.9} & \textbf{47.3} && \textbf{46.1} & \textbf{52.4} && \textbf{22.2} & \textbf{26.7} && \textbf{23.5} & \textbf{29.0} && \textbf{28.5} & \textbf{33.6} && \textbf{28.7} & \textbf{35.5} && \textbf{39.0} & \textbf{44.2}  \\

\bottomrule
\end{tabular}
\caption{A comparison of \textsc{Base} (upper half) and \textsc{Large} (lower half) multilingual encoders on \xlbel.}
%\vspace{-1.5em}
\label{tab:large_models}
\end{table*}

\stitle{Comparison with in-Domain Parallel Data.} While we used general-domain bitexts to cover more resource-poor languages, we are aware that in-domain bitexts exist among several ``mainstream'' languages (\en, \zh, \es, \pt, \fr, \de, \citealt{bawden-etal-2019-findings}).\footnote{\url{http://www.statmt.org/wmt19/biomedical-translation-task.html}} In the future, we plan to also compare with biomedical term/sentence translations on these languages to gain more insights on the impact of domain-shift.

\subsection{Number of Model Parameters}\label{sec:appendix_number_param}
All \textsc{Base} models have $\approx$110M parameters while \textsc{Large} models have $\approx$340M parameters.

\subsection{Hyperparameter Optimisation}\label{sec:hyperparameters}
\Cref{Table:search_space} lists the hyperparameter search space. Note that the chosen hyperparameters yield the overall best performance, but might be suboptimal in any single setting. We used the same random seed across all experiments.

\begin{table}[!ht] %[!htbp]
\small
%\setlength{\tabcolsep}{1pt}
%\vspace*{-0.3cm}
%\renewcommand{\arraystretch}{0.8}
\centering
\begin{tabular}{lr}
\toprule
hyperparameters & search space \\
\midrule
%batch size & \{7500$^\ast$, 15000\footnote{The }\}\\
pretraining learning rate & \texttt{2e-5} \\
pretraining batch size & 512 \\
pretraining training epochs & 1 \\
bitext fine-tuning learning rate  & \texttt{2e-5} \\
bitext fine-tuning batch size & \{64, 128, 256$^\ast$\}\\
bitext fine-tuning epochs & \{1, 2, 3, 4, 5$^\ast$, 10\} \\
\texttt{max\_seq\_length} of tokeniser & 25 \\
$\lambda$ in Online Mining & 0.2\\
$\alpha$ in MS loss (\Cref{eq:loss}) & 2 \\
$\beta$ in MS loss (\Cref{eq:loss}) & 50 \\
$\epsilon$ in MS loss (\Cref{eq:loss}) & 1 \\
\bottomrule
\end{tabular}
\caption{Hyperparameters along with their search grid. $\ast$ marks the values used to obtain the reported results. The hparams without any defied search grid are adopted directly from \citet{liu2020self}.}
\label{Table:search_space}
\end{table}

\subsection{Software and Hardware Dependencies}
All our experiments are implemented using PyTorch 1.7.0 with Automatic Mixed Precision (AMP)\footnote{\url{https://pytorch.org/docs/stable/amp.html}} turned on. The hardware we use is listed in \Cref{Table:hardware}. On this machine, the \textsc{Sap} fine-tuning procedure generally takes 5-10 hours with UMLS data. \textsc{Sap} fine-tuning with translation data takes 10 minutes to 5 hours, depending on the amount of the data. Inference generally takes $<$10 minutes.

\begin{table}[h] %[!htbp]
\small
\setlength{\tabcolsep}{1pt}
%\vspace*{-0.3cm}
%\renewcommand{\arraystretch}{0.8}
\centering
\begin{tabular}{lr}
\toprule
hardware & specification \\
\midrule
 %RAM & CORSAIR\textsuperscript{\textregistered} Vengeance RGB Pro (16 GB) $\times$ 4\\
 RAM & 192 GB \\
 CPU & Intel Xeon W-2255 @3.70GHz, 10-core 20-threads\\
 GPU & NVIDIA GeForce RTX 2080 Ti (11 GB) $\times$ 4\\
\bottomrule
\end{tabular}
\caption{Hardware specifications of the used machine. For \textsc{Large} model training, we use another server with two NVIDIA GeForce RTX 3090 (24 GB).}
\label{Table:hardware}
\end{table}

\end{document}